\documentclass[runningheads]{llncs}
\usepackage{graphicx}
\usepackage{hyperref}       
\usepackage{url}            
\usepackage{booktabs}       
\usepackage{amsfonts}       
\usepackage{nicefrac}       
\usepackage{microtype}      
\usepackage{xcolor}         
\usepackage{amsmath}
\usepackage{graphicx}
\usepackage{multirow} 
\usepackage{array}
\usepackage{marvosym}

\begin{document}
\title{MCIQA-2K: A Multi-Dimensional Dataset and No-Reference Quality Assessment Benchmark for Colorized Images}
\titlerunning{MCIQA-2K}
\author{Yunkai Zhuang\inst{3} \and
Qihang Yan\inst{1,3} \and
Zicheng Zhang\inst{1,2}\(^\dagger\) \and 
Guangtao Zhai\inst{1,2}\(^\dagger\)}
\authorrunning{Y. Zhuang et al.}
%

\institute{
Shanghai Artificial Intelligence Laboratory, Shanghai 200232, China
\and
School of Electronics, Information and Electrical Engineering,  Shanghai Jiao Tong University, Shanghai 200240, China\\
\email{\{zhaiguangtao, zzc1998\}@sjtu.edu.cn}
\and
School of Information Science and Technology, ShanghaiTech University, Shanghai 201210, China\\
\email{\{yanqh2022, zhuangyk2023\}@shanghaitech.edu.cn}
}

\maketitle              

\begin{abstract}
Image colorization is an inherently ill-posed task, since a single grayscale image may correspond to multiple plausible colorized results. Consequently, conventional full-reference image quality assessment (IQA) metrics fail to accurately reflect human perceptual preferences for colorized images. In this paper, we present MCIQA-2K, a large-scale multi-dimensional benchmark specifically designed for no-reference quality assessment of colorized images. We construct a dataset containing 2,000 colorized images generated by five representative colorization models, together with human annotations across three perceptual dimensions: color smearing, semantic color misalignment, and global naturalness. Building upon the proposed benchmark, we further introduce MCIQA, a dedicated multi-branch NR-IQA framework for colorized images. Extensive experiments demonstrate that MCIQA significantly outperforms existing full-reference and no-reference IQA methods on the proposed benchmark, while also exhibiting competitive generalization capability on several widely-used IQA datasets. The dataset and code are publicly available at \href{https://github.com/ARBEZ-ZEBRA/MCIQA}{https://github.com/ARBEZ-ZEBRA/MCIQA}.

\end{abstract}

\section{Introduction}
Image colorization is the task to restore realistic colors to monochrome images, which is a core challenge in low-level computer vision, with vital uses in updating old photographs, revitalizing classic cinema, and creating digital media. While deep learning has pushed the field forward significantly, the task is fundamentally underdetermined because a single grayscale image can logically support a wide variety of authentic color schemes. A piece of clothing, a car, or a landscape, for example, could each be rendered in several different but equally believable ways. Consequently, the concept of a definitive "correct" color version is inherently flawed, meaning that evaluating these algorithms must rely on human visual judgment rather than direct pixel-matching metrics.

Nevertheless, the majority of current colorization frameworks operate within a full-reference (FR) framework, which inherently presumes that only one ideal outcome exists. Standard models like CIC\cite{zhang2016colorful}, BigColor\cite{kim2022bigcolor}, and DDColor\cite{kang2023ddcolor} depend on point-by-point reconstruction errors or similarity measures tied to a solitary target image. This discrepancy between optimization targets and actual human vision frequently forces networks to produce washed-out tones or overly cautious palettes, ultimately restricting the vividness and authenticity of the generated images.

To overcome this constraint, researchers have turned to no-reference (NR) image quality metrics. By utilizing human judgment scores to direct the evaluation process, these techniques harmonize more closely with subjective visual preferences. Nonetheless, current NR frameworks are generally trained on broad quality evaluation benchmarks or datasets tailored for entirely different restoration challenges\cite{ponomarenko2013color,8968750,ahmed2022biq2021}. As a result, their effectiveness in evaluating color-restored media is hindered by an inherent domain mismatch and a shortage of specialized, task-specific training data.

To address this limitation, we present MCIQA-2K, a multi-dimensional benchmark specifically developed for no-reference quality assessment of colorized images. Through extensive empirical analysis, we identify two dominant and recurrent distortion types that strongly influence human perception but are insufficiently modeled by existing generic IQA approaches: (1) color smearing, referring to spatially incoherent or bleeding color regions that disrupt local structural consistency, and (2) semantic color misalignment, where objects are assigned implausible or semantically incorrect colors. 

Based on these observations, MCIQA-2K is constructed to explicitly model perceptual quality from multiple complementary dimensions. We first select 400 high-quality natural images from the COCO Dataset test set\cite{lin2014microsoft}, convert them to grayscale, and subsequently colorize them using five representative colorization models: BigColor\cite{kim2022bigcolor}, CIC\cite{zhang2016colorful}, CT\(^2\)\cite{weng2022ct}, DDColor\cite{kang2023ddcolor}, and GCP-Colorization\cite{wu2021towards}. Each generated image is independently evaluated by 16 human annotators, who provide perceptual quality scores together with dedicated annotations for three dimensions: color smearing, semantic color alignment and global naturalness. In total, MCIQA-2K contains 2,000 annotated colorized images and, to the best of our knowledge, constitutes the largest benchmark for colorized image quality assessment in terms of annotation dimensionality, diversity of generation models, and richness of human perceptual supervision.

To model these distortions in an interpretable and human-aligned manner, we propose MCIQA, a dedicated no-reference IQA framework consists of two specialized assessment branches together with a global naturalness branch. The color smearing branch is designed to capture chromatic inconsistency and abnormal local color diffusion patterns, enabling effective detection of spatial color smearing artifacts. Meanwhile, the semantic color misalignment branch evaluates whether the assigned colors are semantically plausible for the corresponding image content. Both branches are built upon the Qwen3-VL-32B\cite{bai2025qwen3} vision-language model. A global naturalness branch extracts holistic visual representations to capture overall image naturalness and global perceptual quality cues. The outputs from the color smearing branch, semantic color misalignment branch, and global naturalness branch are subsequently integrated through a learnable regression head to predict the final perceptual naturalness quality score.

MCIQA provides a more comprehensive and human-aligned assessment paradigm for colorized images. Furthermore, the introduction of spatially interpretable quality score maps allows the framework not only to predict perceptual quality accurately, but also to explicitly reveal the underlying causes of quality degradation. 

\section{Related Work}
\paragraph{\textbf{Colorization Method}}
Image colorization has progressed rapidly alongside advancements in deep learning, evolving across several distinct architectural paradigms. 

Early approaches relied on Convolutional Neural Networks (CNNs)\cite{zhang2016colorful,Su-CVPR-2020,salmona2022deoldify} to generate colorized images. Subsequent frameworks\cite{wu2021towards,kim2022bigcolor} integrated Generative Adversarial Networks (GANs) to generate more vivid, plausible colors. More recently, transformer-based models\cite{weng2022ct,kang2023ddcolor,du2024multicolor} have demonstrated a superior capacity for capturing long-range dependencies and global contextual cues. Concurrently, diffusion-based models\cite{zabari2023diffusing,li2024coco,liang2025control} have emerged as a dominant paradigm in image generation and restoration. 

Despite these significant architectural milestones, the majority of existing methods continue to be trained and evaluated under rigid, full-reference assumptions. Consequently, they fail to adequately account for or reflect the rich perceptual diversity inherent to valid colorization outcomes.

\begin{figure}[t]
\centering
\includegraphics[width=1\linewidth]{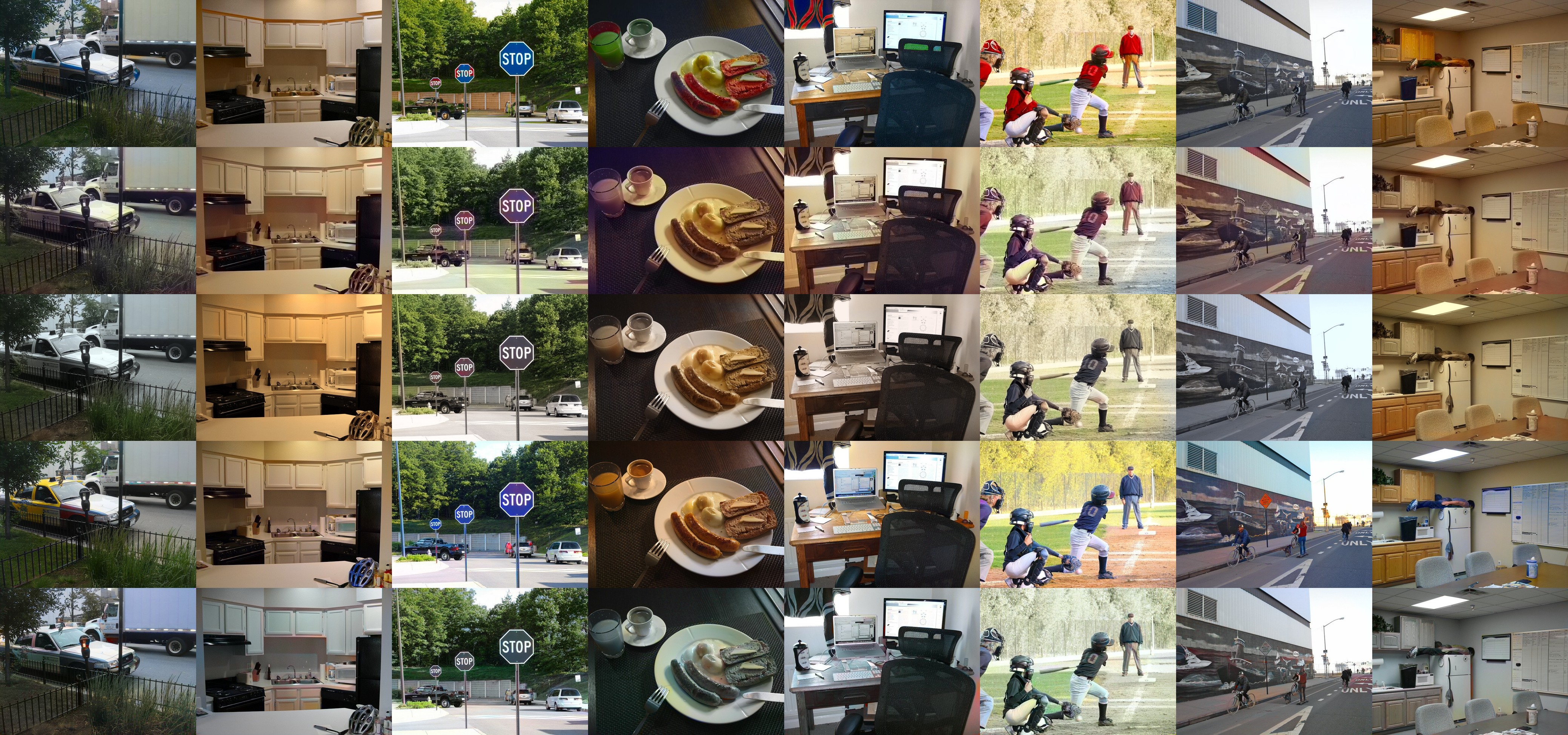}
\caption{
Example image samples of MCIQA-2K dataset.
}
\label{fig:sample}
\end{figure}

\paragraph{\textbf{Dataset}}
The Natural-Color Dataset (NCD)~\cite{anwar2020image} focuses on objects with relatively fixed and canonical color attributes. The Language-based Colorization Benchmark~\cite{li2025language} incorporated textual descriptions to guide the colorization process, promoting semantic consistency between linguistic cues and generated colors. However, these datasets lack human perceptual quality annotations, rendering them unsuitable for training or evaluating Image Quality Assessment (IQA) models. 

CCQ\cite{10929852} leveraged multimodal representations to evaluate perceptual quality. Nonetheless, this dataset was constructed using only two colorization frameworks (DDColor\cite{kang2023ddcolor} and InstColor\cite{Su-CVPR-2020}), which inherently restricts the diversity of the resulting colorization artifacts and stylistic variations.

Beyond colorization-specific benchmarks, a substantial body of work in general no-reference IQA relies on widely adopted datasets \cite{8968750,fang2020cvpr,ahmed2022biq2021}. However, they remain poorly suited for evaluating colorized images because the distortions present in general IQA datasets consist predominantly of low-level degradations and they implicitly operate under the assumption of a single, definitive reference distribution, failing to account for the multi-modal nature of valid colorizations.

\paragraph{\textbf{Image Quality Assessment Method}}
Image quality assessment (IQA) methodologies are broadly classified into full-reference (FR) and no-reference (NR) approaches. FR-IQA metrics, such as PSNR and LPIPS \cite{zhang2018unreasonable}, evaluate fidelity by directly comparing a distorted image against a corresponding ground truth. These methods are ill-suited for ill-posed and fundamentally ambiguous problems like colorization.

Conversely, NR-IQA methods seek to predict perceptual quality in the absence of a reference image. Traditional techniques in this domain rely on natural scene statistics \cite{mittal2012no,mittal2012making}. Meanwhile, contemporary learning-based models \cite{ke2021musiq,chen2024topiq,agnolucci2024quality} leverage large-scale datasets annotated with human opinion scores. These modern approaches are predominantly tailored for generic degradations. As a result, they lack the necessary sensitivity to detect colorization-specific artifacts, ultimately limiting their efficacy in this specialized domain.

\section{Dataset Construction}
\subsection{Data Selection}
To guarantee that our benchmark faithfully reflects real-world practical performance, we curate 400 high-quality grayscale images from the COCO test2017 dataset\cite{lin2014microsoft}. The selected samples cover diverse semantic categories, varying scene complexities and diverse lighting scenarios, featuring abundant visual variations and distinct object structures to facilitate reliable subsequent image colorization.

To fully encompass diverse colorization characteristics and typical visual artifacts, we adopt five mainstream colorization models with distinct network architectures: CIC\cite{zhang2016colorful}(CNN), GCP-Colorization\cite{wu2021towards}(GAN), BigColor\cite{kim2022bigcolor}(GAN), CT\(^2\)\cite{weng2022ct}(Transformer) and DDColor\cite{kang2023ddcolor}(Transformer). In total, we obtain 2000 colorized images. These models differ significantly in color restoration accuracy, generation style and error manifestation patterns, producing results ranging from muted, low-saturation color outputs to visually vibrant yet semantically unreasonable colorization outcomes. Such multi-model diversity allows our dataset to closely simulate the performance variance of practical colorization algorithms, and comprehensively include prevalent visual defects such as color confusion and semantic color misalignment. Several image samples are shown in Fig\ref{fig:sample}

\subsection{Human Annotation}
To enhance the reliability of Mean Opinion Score (MOS) evaluations, we recruited a diverse group of human annotators for subjective scoring. In total, 16 qualified raters including 10 males and 6 females took part in this study, all of whom possessed normal or corrected visual acuity and confirmed standard color vision capability. More details can be found in the supplementary material.

The annotation instruction is: "Color smearing refers to spatially incoherent or bleeding color regions that disrupt local structural consistency. Semantic color misalignment means that objects are assigned implausible or semantically incorrect colors. Global naturalness is the holistic visual representations to capture overall image naturalness. Given the two images below, the image on the left is the grayscale version of the colorized image on the right. You may refer to the grayscale image as guidance, and rate the colorized image from the dimensions of color smearing, semantic color misalignment and global naturalness on a scale of 1 to 5, where 1 indicates the lowest quality and 5 indicates the highest quality.". 

To further verify the consistency of subjective annotations, we calculated inter-rater reliability via Krippendorff’s Alpha based on interval scales. The obtained scores reached 0.7794 for color smearing, 0.7625 for semantic color misalignment, and 0.8163 for global visual naturalness. Such favorable agreement values demonstrate high consensus among all raters and solidly validate the credibility of the acquired subjective perceptual scores.

\subsection{Normalization}
To account for inter-subject variability, raw ratings were normalized on a per-scorer basis. Specifically, the raw rating $r_{ijk}$ assigned by the $i$-th scorer to the $j$-th image along the $k$-th dimension is converted into a Z-score $z_{ijk}$ as follows:
\begin{equation}
z_{ijk} = \frac{r_{ijk} - \mu_{ik}}{\sigma_{ik}}
\end{equation}
\begin{equation}
\mu_{ik} = \frac{1}{N_{ik}}\sum_{j=1}^{N_{ik}}r_{ijk}
\end{equation}
\begin{equation}
\sigma_{ik} = \sqrt{\frac{1}{N_{ik}-1}\sum_{j=1}^{N_{ik}}(r_{ijk}-\mu_{ik})^2}
\end{equation}
where $N_{ik}$ denotes the number of images rated by scorer $i$ along the k-th dimension. The final MOS for image $j$ in the k-th dimension is computed by averaging normalized scores across all scorers:
\begin{equation}
\text{MOS}_{jk} = \frac{1}{M}\sum_{i=1}^{M} z_{ijk}
\end{equation}
where $M$ is the total number of scorers.

\begin{figure}[htb]
  \centering
  \centerline{\includegraphics[width=1\linewidth]{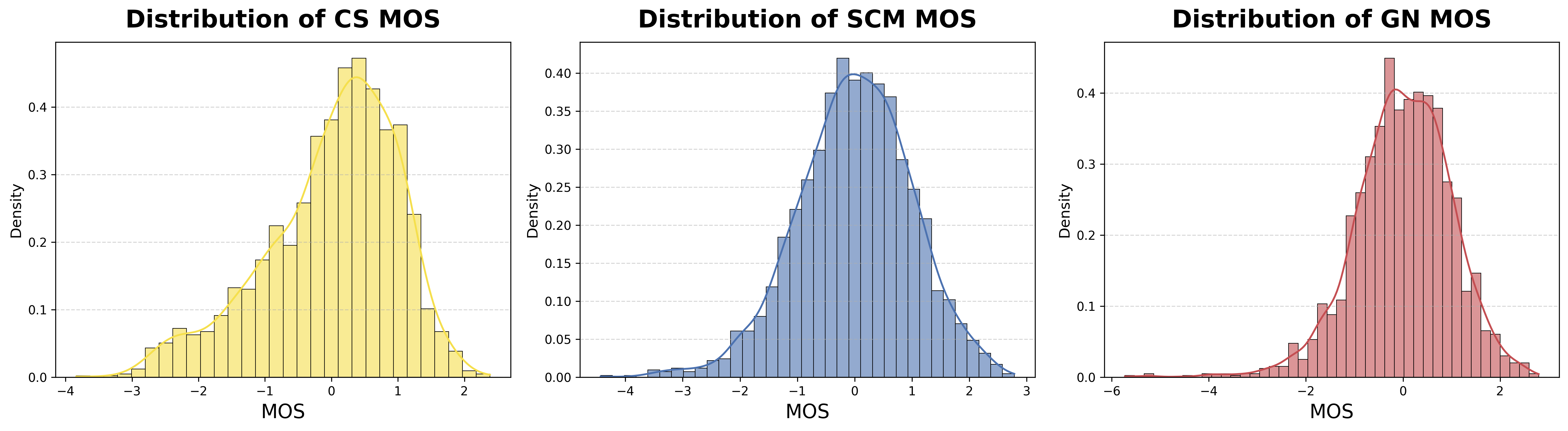}}
\caption{Illustration of MOS distribution across all samples.}
\label{fig:MOS}
\end{figure}

To further analyze the statistical properties of the collected annotations, we visualize the distribution of Mean Opinion Scores (MOS) across all samples in Fig.~\ref{fig:MOS}. As illustrated, the scores in all three dimensions consistently exhibit a near-normal (Gaussian) distribution, validating the consensus and consistency among the human annotators.

To verify the independence of the proposed dimensions, we evaluated their inter-dimension correlations. The SRCC / PLCC are $0.4700 / 0.4968$ between color smearing and semantic color misalignment, $0.5377 / 0.5536$ between global naturalness and color smearing, and $0.4610 / 0.5474$ between global naturalness and semantic color misalignment. These relatively low correlation metrics clearly underscore the significant discrepancies among the three dimensions, confirming that each dimension successfully characterizes distinct perceptual information without severe information redundancy.

\section{Methodology}
\subsection{Problem Analysis}
Despite the remarkable advancements in contemporary image colorization techniques, their generated results still frequently suffer from distinct perceptual artifacts, which cannot be effectively quantified by conventional image quality assessment (IQA) metrics. Based on extensive empirical analysis on various colorization outputs, we summarize two prevalent and typical failure patterns.

The first artifact is color smearing, referring to spatially inhomogeneous color overflow that destroys local visual coherence. This defect mainly emerges near object edges and texture-rich areas, where colors diffuse beyond reasonable semantic boundaries and produce unnatural visual transitions.

The second type is semantic color misalignment, in which unreasonable or factually incorrect hues are assigned to target objects. Such inappropriate color matching violates real-world semantic logic and severely degrades overall visual authenticity.

\begin{figure}[htb]
  \centering
  \centerline{\includegraphics[width=1\linewidth]{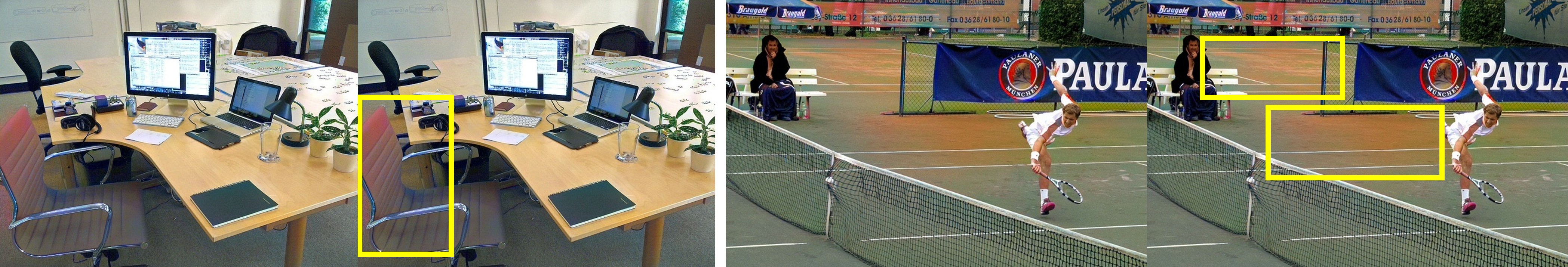}}
\caption{Color smearing, where colors bleed across object boundaries, resulting in spatial inconsistency.}
\label{fig:CS}
\end{figure}

\begin{figure}[htb]
  \centering
  \centerline{\includegraphics[width=1\linewidth]{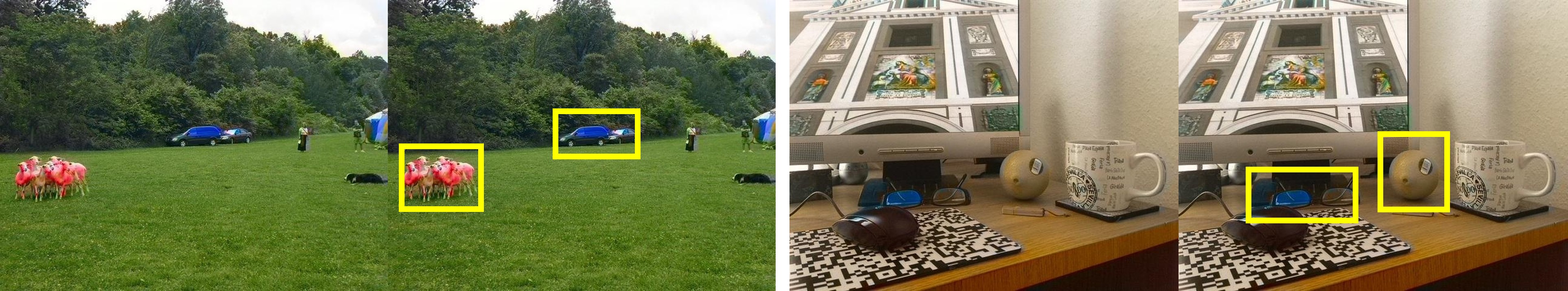}}
\caption{Semantic color misalignment, where objects are assigned implausible colors that contradict real-world semantics.}
\label{fig:SCM}
\end{figure}

Typical cases of the above two distortion categories are presented in Fig.~\ref{fig:CS} and Fig.~\ref{fig:SCM}, which intuitively demonstrate their unique visual manifestations.

\subsection{Architecture}
\begin{figure}[htb]
  \centering
  \centerline{\includegraphics[width=1\linewidth]{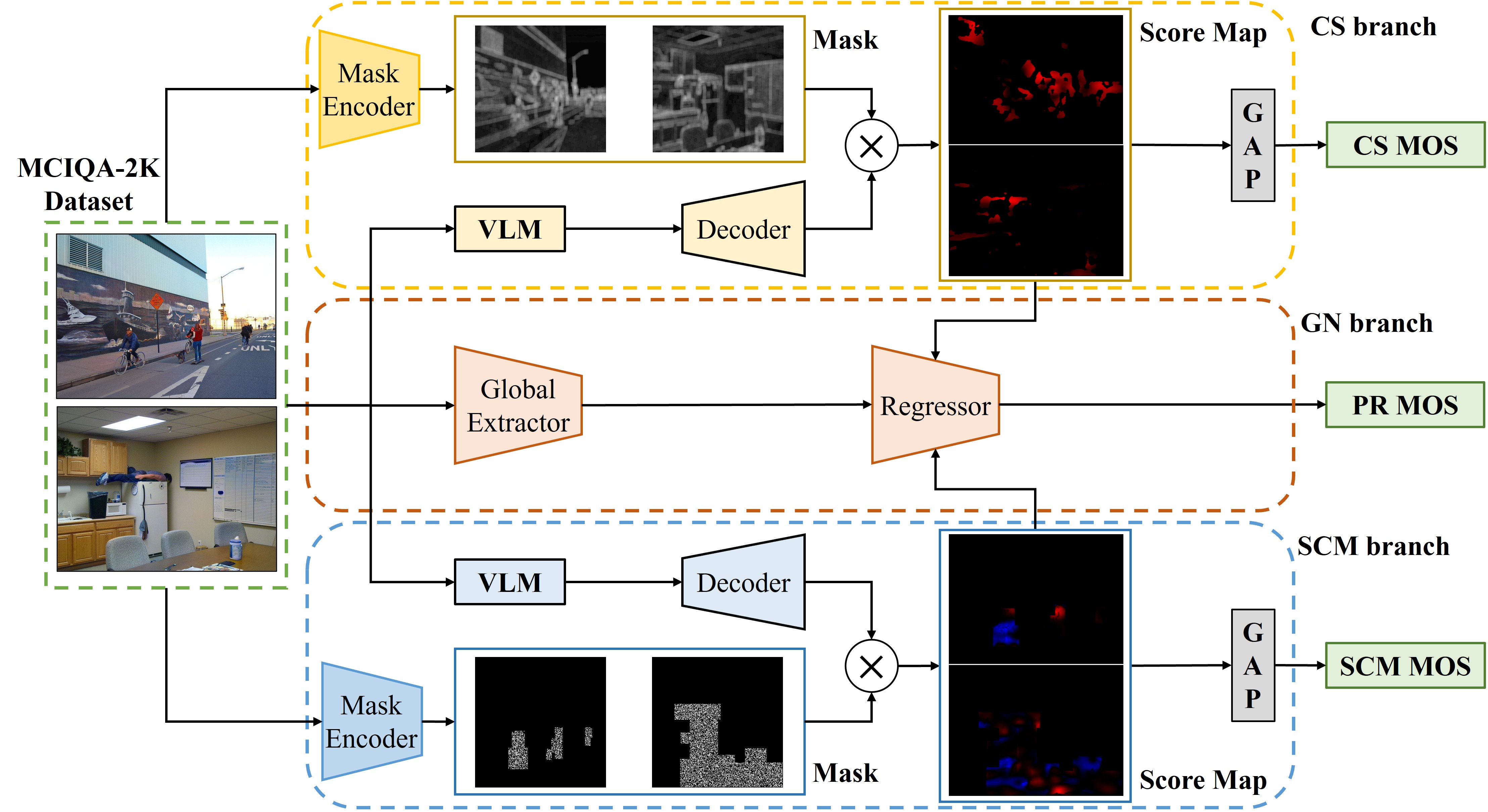}}
\caption{Illustration of the architecture of MCIQA.}
\label{fig:architecture}
\end{figure}

To address the perceptual challenges discussed above, we propose \textbf{MCIQA}, a multi-branch no-reference IQA framework specifically designed for colorized images. As illustrated in Fig.~\ref{fig:architecture}, MCIQA integrates three complementary components: a color smearing (CS) branch, a semantic color misalignment (SCM) branch, and a global naturalness (GN) branch. 

The architectures of the CS and SCM branches are structurally similar. Given an input colorized image, a fine-tuned vision-language model, Qwen \textbf{Qwen3-VL-32B}\cite{bai2025qwen3} is first employed to extract dimension-aware visual representations. Benefiting from the large-scale visual-semantic knowledge learned during pre-training, the VLM is capable of identifying semantically meaningful regions and recognizing visually abnormal color patterns that are strongly correlated with human perceptual judgments. To better align the extracted features with the corresponding perceptual dimension, only the visual encoder of the VLM is fine-tuned during training. The VLM outputs a two-dimensional feature representation that emphasizes distortion-relevant spatial information for the target dimension.

Subsequently, the generated feature maps are fed into an attention-enhanced CNN module to further capture critical local structures and perceptually important regions. The attention mechanism enables the network to selectively emphasize informative distortion patterns while suppressing irrelevant content, thereby improving sensitivity to subtle colorization artifacts.

Meanwhile, a ViT-based mask generation network takes the original colorized image as input and predicts a spatial importance mask. The generated mask aims to localize regions that are highly correlated with human perceptual judgments for the corresponding quality dimension. The mask is then multiplied element-wise with the refined CNN feature maps to produce a dimension-specific score map: \begin{equation}S_{map} = M \times F\end{equation}, where M denotes the predicted spatial mask and F represents the CNN-refined feature representation.

Afterward, a simple Global Average Pooling (GAP) operation is applied to the score map to obtain the final quality prediction score for the corresponding dimension. The predicted score is supervised using the subjective annotation score from the proposed dataset. Importantly, the use of GAP establishes a strong direct correlation between the two-dimensional score map and the final scalar quality score, which significantly improves the interpretability and readability of the learned score maps. As a result, regions with severe perceptual degradation contribute more prominently to the final prediction.

Different from the CS and SCM branches that focus on localized distortion perception, the GN branch aims to evaluate the holistic visual naturalness of the colorized image. Specifically, a ResNet-50-based\cite{7780459} global feature extractor is first employed to capture high-level semantic and appearance information from the input image. The extracted global representation is then fused with the score maps generated by the CS and SCM branches. Finally, a lightweight regressor predicts the global naturalness score by jointly considering global image statistics and localized distortion-aware cues. This design allows the GN branch to leverage both semantic consistency and local artifact distributions, leading to more reliable naturalness assessment for colorized images.

To optimize the quality prediction network, we employ a composite objective function, which jointly considers score regression accuracy and relative ranking consistency. Specifically, the loss function consists of a mean squared error (MSE) term and a pairwise ranking regularization term: \begin{equation}\mathcal{L} = \mathcal{L}_{MSE} + \lambda\mathcal{L}_{rank}\end{equation}\begin{equation}\mathcal{L}_{rank} = \frac{1}{N^2}\sum_{i,j}\max(p_j-p_i)\text{sign}(t_i-t_j)\end{equation}, where $\lambda$ denotes the weighting coefficient balancing the two objectives, N represents the size of dataset, $p_i$, $p_j$ represent predicted quality scores and $t_i$, $t_j$ represent ground-truth quality scores.

The ranking term penalizes inconsistent relative ordering between predicted scores and ground-truth annotations. This design encourages the model to maintain correct perceptual ranking relationships across different images.

By combining regression supervision and ranking-aware optimization, the loss improves both prediction precision and perceptual consistency.

\section{Experiment} 
\subsection{Image Quality Assessment Methods} 
The selected methods can be classified into three groups. Full-reference methods: PSNR, SSIM\cite{wang2004image}, LPIPS\cite{zhang2018unreasonable} and DISTS\cite{ding2020image} compare the inference results with the reference images. No-reference handcrafted methods: BRISQUE\cite{mittal2012no} and NIQE\cite{mittal2012making} extract handcrafted features from the inference results. No-reference deep learning methods: CNN-IQA\cite{kang2014convolutional}, DBCNN\cite{zhang2018blind}, WaDIQaM\cite{bosse2017deep}, MUSIQ\cite{ke2021musiq}, CLIP-IQA\cite{wang2023exploring}, CLIP-IQA+\cite{wang2023exploring} and QualiClip\cite{agnolucci2024quality} use deep neural networks trained from labeled data to characterize quality-aware information of the inference results.

\begin{table}[htbp]
    \centering
    \caption{Performance comparison of different metrics on the MCIQA-2K dataset based on the value of SRCC and PLCC. The best results are highlighted in \textcolor{red}{red}. The second best results are highlighted in \textcolor{blue}{blue}. The reference images of all FR metrics are the original RGB images from COCO test set\cite{lin2014microsoft}. $\uparrow$ means 1 is the best and -1 is the worst. $\downarrow$ means -1 is the best and 1 is the worst. The value of each IQA on each dataset is the average result of 3 experiments with different random seeds. }
    \label{tab:performance_comparison}
    \begin{tabular}{@{}lcccccc@{}}
        \toprule
        \multirow{2}{*}{Metric} & \multicolumn{2}{c}{MCIQA-2K-CS} & \multicolumn{2}{c}{MCIQA-2K-SCM} & \multicolumn{2}{c}{MCIQA-2K-GN} \\
        \cmidrule(lr){2-3} \cmidrule(lr){4-5} \cmidrule(lr){6-7} & SRCC & PLCC & SRCC & PLCC & SRCC & PLCC \\
        \midrule
        {PSNR}$\uparrow$ & \textcolor{blue}{0.6131} & \textcolor{blue}{0.6229} & 0.3710 & 0.4037 & 0.4644 & 0.4797 \\
        {SSIM}$\uparrow$& -0.0268 & 0.0136 & -0.1332 & -0.0682 & 0.0639 & 0.0240 \\
        {LPIPS}$\downarrow$ & -0.5559 & -0.5735 & \textcolor{blue}{-0.4847} & \textcolor{blue}{-0.5205} & -0.5122 & -0.5164 \\
        {DISTS}$\downarrow$ & -0.3527 & -0.3749 & -0.2686 & -0.2938 & -0.3508 & -0.3650 \\
        \midrule
        {BRISQUE}$\downarrow$ & 0.0062 & 0.0200 & -0.0101 & 0.0248 & 0.0563 & 0.0591 \\
        {NIQE}$\downarrow$ & -0.0441 & -0.0471 & -0.0391 & -0.0446 & -0.1544 & -0.1720 \\
        \midrule
        {CNN-IQA}$\uparrow$ & 0.0338 & 0.0472 & 0.0944 & 0.1875 & 0.0970 & 0.1843 \\
        {DBCNN}$\uparrow$ & 0.2196 & 0.2319 & 0.2123 & 0.3235 & 0.3651 & 0.4469 \\
        {WaDIQaM}$\uparrow$ & 0.1629 & 0.2018 & 0.1396 & 0.2514 & 0.2210 & 0.3315 \\
        {MUSIQ}$\uparrow$ & 0.1422 & 0.1861 & 0.0978 & 0.2184 & 0.2872 & 0.4042 \\
        {CLIP-IQA}$\uparrow$ & 0.2468 & 0.2614 & 0.3223 & 0.3879 & 0.5619 & 0.5968 \\
        {CLIP-IQA+}$\uparrow$ & 0.2327 & 0.2445 & 0.2939 & 0.3689 & 0.4019 & 0.4749 \\
        {QualiClip}$\uparrow$ & 0.3199 & 0.3305 & 0.3961 & 0.4964 & \textcolor{blue}{0.7126} & \textcolor{blue}{0.7545} \\
        \midrule
        {\textbf{MCIQA-CS}}$\uparrow$ & \textcolor{red}{0.8274} & \textcolor{red}{0.8213} & 0.4660 & 0.4642 & 0.5045 & 0.5011 \\
        {\textbf{MCIQA-SCM}}$\uparrow$ & 0.4597 & 0.4817 & \textcolor{red}{0.8902} & \textcolor{red}{0.8408} & 0.4522 & 0.5648 \\
        {\textbf{MCIQA-GN}}$\uparrow$ & 0.5059 & 0.5193 & 0.3938 & 0.4749 & \textcolor{red}{0.8511} & \textcolor{red}{0.8571} \\
        \bottomrule
    \end{tabular}
\end{table}
\begin{table}[htbp]
    \centering
    \caption{Performance comparison of different metrics on the different datasets based on the value of SRCC and PLCC. The value of each IQA on each dataset is the average result of 3 experiments with different random seeds. }
    \label{tab:performance_comparison_}
    \begin{tabular}{@{}lcccccccc@{}}
        \toprule
        \multirow{2}{*}{Metric} & \multicolumn{2}{c}{KONIQ-10K} & \multicolumn{2}{c}{BIQ2021} & \multicolumn{2}{c}{AGIQA-1K} & \multicolumn{2}{c}{AGIQA-3K} \\
        \cmidrule(lr){2-3} \cmidrule(lr){4-5} \cmidrule(lr){6-7} \cmidrule(lr){8-9} & SRCC & PLCC & SRCC & PLCC & SRCC & PLCC & SRCC & PLCC \\
        \midrule
        \small{BRISQUE}$\downarrow$ & -0.1797 & -0.1603 & -0.4502 & -0.4674 & -0.4958 & -0.5387 & -0.4728 & -0.5007 \\
        \small{NIQE}$\downarrow$ & -0.3107 & -0.2330 & -0.4366 & -0.4735 & -0.5594 & -0.5806 & -0.5075 & -0.4847 \\
        \midrule
        \small{CNN-IQA}$\uparrow$ & 0.7199 & 0.7667 & 0.5067 & 0.5531 & 0.1461 & 0.3136 & 0.6006 & \textcolor{red}{0.7724} \\
        \small{DBCNN}$\uparrow$ & \textcolor{red}{0.9021} & \textcolor{red}{0.9198} & \textcolor{blue}{0.6743} & 0.7270 & 0.4502 & 0.4620 & 0.6398 & 0.7483 \\
        \small{WaDIQaM}$\uparrow$ & 0.7824 & 0.8188 & 0.5797 & 0.6407 & 0.2675 & 0.3609 & 0.5705 & 0.7254 \\
        \small{MUSIQ}$\uparrow$ & \textcolor{blue}{0.8511} & \textcolor{blue}{0.8706} & \textcolor{red}{0.7287} & \textcolor{red}{0.7679} & 0.4246 & 0.5461 & 0.6207 & 0.7157 \\
        \small{CLIP-IQA}$\uparrow$ & 0.5502 & 0.5962 & 0.5061 & 0.5662 & 0.3307 & 0.3948 & 0.6561 & 0.7306 \\
        \small{CLIP-IQA+}$\uparrow$ & 0.7893 & 0.8155 & 0.6714 & 0.7199 & 0.4359 & 0.5310 & \textcolor{red}{0.6862} & \textcolor{blue}{0.7514} \\
        \small{QualiClip}$\uparrow$ & 0.6238 & 0.6667 & 0.6730 & \textcolor{blue}{0.7341} & \textcolor{blue}{0.6282} & \textcolor{red}{0.6533} & \textcolor{blue}{0.6634} & 0.7168 \\
        \midrule
        \small{\textbf{MCIQA-CS}}$\uparrow$ & 0.5018 & 0.5867 & 0.4482 & 0.4592 & 0.5101 & 0.5090 & 0.5741 & 0.5756 \\
        \small{\textbf{MCIQA-SCM}}$\uparrow$ & 0.4048 & 0.4380 & 0.3676 & 0.4231 & 0.4964 & 0.5238 & 0.4495 & 0.4870 \\
        \small{\textbf{MCIQA-GN}}$\uparrow$ & 0.7502 & 0.7987 & 0.7071 & 0.7353 & \textcolor{red}{0.6373} & \textcolor{blue}{0.6514} & 0.6534 & 0.7216 \\
        \bottomrule
    \end{tabular}
\end{table}

\subsection{Experimental Datasets}
To thoroughly assess the effectiveness and generalization capability of our proposed method, we evaluate it across several widely adopted perceptual image quality assessment (IQA) benchmarks: KONIQ-10K \cite{8968750}, BIQ2021 \cite{ahmed2022biq2021}, AGIQA-1K \cite{10222021}, and AGIQA-3K \cite{li2023agiqa}.

KONIQ-10k \cite{8968750} is a massive, in-the-wild IQA dataset that comprises diverse authentic distortions found in real-world imagery, paired with human-annotated mean opinion scores (MOS). BIQ2021 \cite{ahmed2022biq2021} builds upon this foundation by introducing a wider array of realistic distortions and complex evaluation scenarios. AGIQA-1K \cite{10222021} and AGIQA-3K \cite{li2023agiqa} represent recent benchmarks dedicated to AI-generated content, where perceptual quality is governed not just by low-level degradations, but crucially by high-level semantic plausibility and overall visual realism.

\subsection{Evaluation Criteria}
To quantitatively evaluate the performance of our proposed MCIQA against existing metrics, we employ two standard evaluation criteria in the field of Image Quality Assessment (IQA): Spearman Rank Correlation Coefficient (SRCC) and Pearson Linear Correlation Coefficient (PLCC). While SRCC assesses prediction monotonicity based on rank order, PLCC measures the linear accuracy of the quality predictions. Both metrics effectively evaluate the alignment between the objective scores generated by the algorithms and the subjective Mean Opinion Scores (MOS) assigned by human subjects.

\subsection{Experimental Setup}
The MCIQA-2K dataset is randomly split into training and testing sets with an 80/20 ratio, ensuring that colorized images derived from the same grayscale source are strictly assigned to the same split to prevent data leakage.

Our proposed MCIQA model is trained using the Adam optimizer with a batch size of 16 across 6 NVIDIA GeForce RTX 5090 GPUs. The learning rates are empirically set as follows: 1e-5 for the vision-language model (VLM), mask encoder, and decoder in the color smearing (CS) and semantic color misalignment (SCM) branches; and 1e-4 for the global extractor and regressor in the global naturalness (GN) branch. Specifically, the CS branch is trained for 400 epochs (approx. 16 hours), the SCM branch for 300 epochs (approx. 13 hours), and the GN branch for 160 epochs (approx. 8 hours).

\subsection{Experimental Results}
The quantitative evaluation results on the proposed MCIQA-2K dataset and several widely-used IQA benchmarks are presented in Table~\ref{tab:performance_comparison} and Table~\ref{tab:performance_comparison_}. Overall, the proposed MCIQA framework consistently achieves superior performance on the corresponding distortion dimensions, demonstrating its effectiveness in modeling colorization-specific perceptual degradations.

Full-reference IQA metrics exhibit limited capability. Although PSNR achieves moderate performance on the color smearing subset due to its sensitivity to low-level pixel deviations, it performs poorly on other subsets, indicating that pixel fidelity alone cannot adequately characterize perceptual colorization quality. Similarly, SSIM\cite{wang2004image}, LPIPS\cite{zhang2018unreasonable} and DISTS\cite{ding2020image}, still struggle to model semantic color consistency and global realism.

Conventional no-reference IQA methods, including BRISQUE\cite{mittal2012no} and NIQE\cite{mittal2012making}, also fail to generalize to colorization artifacts because they are  designed for traditional distortions such as blur, noise, and compression artifacts.

Deep learning based NR-IQA approaches achieve improved performance compared. Among them, CLIP-based methods, including CLIP-IQA\cite{wang2023exploring}, CLIP-IQA+\cite{wang2023exploring} and QualiClip\cite{agnolucci2024quality}, demonstrate relatively strong performance, particularly on the GN subset, owing to their ability to exploit high-level semantic representations from vision-language models. However, these general-purpose IQA models still show limited capability in handling color smearing and semantic color inconsistency, as they are not explicitly designed for colorization-specific distortions.

In contrast, the proposed MCIQA branches achieve the best performance on their corresponding perceptual dimensions. These results validate that different colorization artifacts require specialized perceptual modeling strategies, and further justify the necessity of the proposed multi-branch architecture.

Table~\ref{tab:performance_comparison_} further evaluates the generalization capability of the proposed method on existing IQA benchmarks. Although MCIQA is specifically designed for colorized image assessment, the GN branch demonstrates competitive cross-dataset performance. In particular, MCIQA-GN achieves the best SRCC on AGIQA-1K (0.6373) and competitive results on BIQ2021 and AGIQA-3K, surpassing or matching several state-of-the-art NR-IQA methods. These results suggest that the global naturalness modeling strategy learned from colorized images can generalize effectively to broader perceptual quality assessment scenarios, especially for AI-generated content. Meanwhile, the CS and SCM branches show relatively lower performance on conventional IQA datasets, which is expected since these datasets do not explicitly contain the specialized colorization distortions targeted by the proposed framework.

\section{Conclusion} 
In this paper, we presented MCIQA-2K, a multi-dimensional benchmark specifically designed for no-reference quality assessment of colorized images. The dataset contains 2,000 colorized images generated by five representative colorization frameworks and is annotated with three dimensional scores collected from 16 annotators.

We further introduced MCIQA, a dedicated multi-branch NR-IQA framework for colorized images. By jointly modeling local chromatic consistency, semantic color plausibility, and holistic visual naturalness, the proposed method achieves substantially improved perceptual quality prediction performance. Extensive experimental results and ablation studies demonstrate the effectiveness of the proposed specialized branches and validate the necessity of dimension-aware perceptual modeling for colorization assessment. Furthermore, the learned global naturalness branch exhibits encouraging generalization ability on multiple existing IQA benchmarks.

Overall, this work establishes a more human-aligned and interpretable evaluation paradigm for image colorization quality assessment. We hope that the proposed dataset and framework can facilitate future research on perceptual evaluation, controllable colorization, and quality-aware optimization for generative image restoration models.

\section{Acknowledgement}
This work was supported by New Generation Artificial Intelligence-National Science and Technology Major Project (2025ZD0124104) in collaboration with Shanghai Artificial Intelligence Laboratory.\\

This work was supported by the Shanghai Municipal Special Program for Basic Research on General AI Foundation Models (Grant No. 2025SHZDZX025D09), in collaboration with Shanghai Artificial Intelligence Laboratory.\\

The authors gratefully acknowledge the GPU computing resources provided by the VSP laboratory of Professor Lou Xin from ShanghaiTech University.\\

%
%
%

%
%
%
%

\newpage
\bibliographystyle{splncs04} 
\bibliography{refs}

@article{mittal2012making,
  title={Making a “completely blind” image quality analyzer},
  author={Mittal, Anish and Soundararajan, Rajiv and Bovik, Alan C},
  journal={IEEE Signal processing letters},
  volume={20},
  number={3},
  pages={209--212},
  year={2012},
  publisher={IEEE}
}

@article{mittal2012no,
  title={No-reference image quality assessment in the spatial domain},
  author={Mittal, Anish and Moorthy, Anush Krishna and Bovik, Alan Conrad},
  journal={IEEE Transactions on image processing},
  volume={21},
  number={12},
  pages={4695--4708},
  year={2012},
  publisher={IEEE}
}

@inproceedings{ke2021musiq,
  title={Musiq: Multi-scale image quality transformer},
  author={Ke, Junjie and Wang, Qifei and Wang, Yilin and Milanfar, Peyman and Yang, Feng},
  booktitle={Proceedings of the IEEE/CVF international conference on computer vision},
  pages={5148--5157},
  year={2021}
}

@article{zhang2018blind,
  title={Blind image quality assessment using a deep bilinear convolutional neural network},
  author={Zhang, Weixia and Ma, Kede and Yan, Jia and Deng, Dexiang and Wang, Zhou},
  journal={IEEE Transactions on Circuits and Systems for Video Technology},
  volume={30},
  number={1},
  pages={36--47},
  year={2018},
  publisher={IEEE}
}

@article{agnolucci2024quality,
  title={Quality-aware image-text alignment for opinion-unaware image quality assessment},
  author={Agnolucci, Lorenzo and Galteri, Leonardo and Bertini, Marco},
  journal={arXiv preprint arXiv:2403.11176},
  year={2024}
}

@inproceedings{wang2023exploring,
  title={Exploring clip for assessing the look and feel of images},
  author={Wang, Jianyi and Chan, Kelvin CK and Loy, Chen Change},
  booktitle={Proceedings of the AAAI conference on artificial intelligence},
  volume={37},
  number={2},
  pages={2555--2563},
  year={2023}
}

@article{bosse2017deep,
  title={Deep neural networks for no-reference and full-reference image quality assessment},
  author={Bosse, Sebastian and Maniry, Dominique and M{\"u}ller, Klaus-Robert and Wiegand, Thomas and Samek, Wojciech},
  journal={IEEE Transactions on image processing},
  volume={27},
  number={1},
  pages={206--219},
  year={2017},
  publisher={IEEE}
}

@inproceedings{kang2014convolutional,
  title={Convolutional neural networks for no-reference image quality assessment},
  author={Kang, Le and Ye, Peng and Li, Yi and Doermann, David},
  booktitle={Proceedings of the IEEE conference on computer vision and pattern recognition},
  pages={1733--1740},
  year={2014}
}

@article{wang2004image,
  title={Image quality assessment: from error visibility to structural similarity},
  author={Wang, Zhou and Bovik, Alan C and Sheikh, Hamid R and Simoncelli, Eero P},
  journal={IEEE transactions on image processing},
  volume={13},
  number={4},
  pages={600--612},
  year={2004},
  publisher={IEEE}
}

@inproceedings{zhang2018unreasonable,
  title={The unreasonable effectiveness of deep features as a perceptual metric},
  author={Zhang, Richard and Isola, Phillip and Efros, Alexei A and Shechtman, Eli and Wang, Oliver},
  booktitle={Proceedings of the IEEE conference on computer vision and pattern recognition},
  pages={586--595},
  year={2018}
}

@article{ding2020image,
  title={Image quality assessment: Unifying structure and texture similarity},
  author={Ding, Keyan and Ma, Kede and Wang, Shiqi and Simoncelli, Eero P},
  journal={IEEE transactions on pattern analysis and machine intelligence},
  volume={44},
  number={5},
  pages={2567--2581},
  year={2020},
  publisher={IEEE}
}

@inproceedings{zhang2016colorful,
  title={Colorful image colorization},
  author={Zhang, Richard and Isola, Phillip and Efros, Alexei A},
  booktitle={European conference on computer vision},
  pages={649--666},
  year={2016},
  organization={Springer}
}

@inproceedings{wu2021towards,
  title={Towards vivid and diverse image colorization with generative color prior},
  author={Wu, Yanze and Wang, Xintao and Li, Yu and Zhang, Honglun and Zhao, Xun and Shan, Ying},
  booktitle={Proceedings of the IEEE/CVF international conference on computer vision},
  pages={14377--14386},
  year={2021}
}

@inproceedings{kim2022bigcolor,
  title={Bigcolor: Colorization using a generative color prior for natural images},
  author={Kim, Geonung and Kang, Kyoungkook and Kim, Seongtae and Lee, Hwayoon and Kim, Sehoon and Kim, Jonghyun and Baek, Seung-Hwan and Cho, Sunghyun},
  booktitle={European Conference on Computer Vision},
  pages={350--366},
  year={2022},
  organization={Springer}
}

@inproceedings{weng2022ct,
  title={CT 2: Colorization transformer via color tokens},
  author={Weng, Shuchen and Sun, Jimeng and Li, Yu and Li, Si and Shi, Boxin},
  booktitle={European Conference on Computer Vision},
  pages={1--16},
  year={2022},
  organization={Springer}
}

@inproceedings{kang2023ddcolor,
  title={Ddcolor: Towards photo-realistic image colorization via dual decoders},
  author={Kang, Xiaoyang and Yang, Tao and Ouyang, Wenqi and Ren, Peiran and Li, Lingzhi and Xie, Xuansong},
  booktitle={Proceedings of the IEEE/CVF International Conference on Computer Vision},
  pages={328--338},
  year={2023}
}

@inproceedings{lin2014microsoft,
  title={Microsoft coco: Common objects in context},
  author={Lin, Tsung-Yi and Maire, Michael and Belongie, Serge and Hays, James and Perona, Pietro and Ramanan, Deva and Doll{\'a}r, Piotr and Zitnick, C Lawrence},
  booktitle={European conference on computer vision},
  pages={740--755},
  year={2014},
  organization={Springer}
}

@article{anwar2020image,
  title={Image colorization: A survey and dataset},
  author={Anwar, Saeed and Tahir, Muhammad and Li, Chongyi and Mian, Ajmal and Khan, Fahad Shahbaz and Muzaffar, Abdul Wahab},
  journal={arXiv preprint arXiv:2008.10774},
  year={2020}
}

@article{li2025language,
  title={Language-based Image Colorization: A Benchmark and Beyond},
  author={Li, Yifan and Yang, Shuai and Liu, Jiaying},
  journal={arXiv preprint arXiv:2503.14974},
  year={2025}
}

@INPROCEEDINGS{10929852,
  author={Shimizu, Shunta and Ishikawa, Hiroshi},
  booktitle={2025 IEEE International Conference on Consumer Electronics (ICCE)}, 
  title={Colorization Quality Assessment with CLIP}, 
  year={2025},
  volume={},
  number={},
  pages={1-6},
  doi={10.1109/ICCE63647.2025.10929852}
}

@inproceedings{Su-CVPR-2020,
  author = {Su, Jheng-Wei and Chu, Hung-Kuo and Huang, Jia-Bin},
  title = {Instance-aware Image Colorization},
  booktitle = {IEEE Conference on Computer Vision and Pattern Recognition (CVPR)},
  year = {2020}
}

@ARTICLE{8968750,
  author={Hosu, Vlad and Lin, Hanhe and Sziranyi, Tamas and Saupe, Dietmar},
  journal={IEEE Transactions on Image Processing}, 
  title={KonIQ-10k: An Ecologically Valid Database for Deep Learning of Blind Image Quality Assessment}, 
  year={2020},
  volume={29},
  number={},
  pages={4041-4056},
  doi={10.1109/TIP.2020.2967829}
  }

@inproceedings{ponomarenko2013color,
  title={Color image database TID2013: Peculiarities and preliminary results},
  author={Ponomarenko, Nikolay and Ieremeiev, Oleg and Lukin, Vladimir and Egiazarian, Karen and Jin, Lina and Astola, Jaakko and Vozel, Benoit and Chehdi, Kacem and Carli, Marco and Battisti, Federica and others},
  booktitle={European workshop on visual information processing (EUVIP)},
  pages={106--111},
  year={2013},
  organization={IEEE}
}

@article{ahmed2022biq2021,
  title={BIQ2021: a large-scale blind image quality assessment database},
  author={Ahmed, Nisar and Asif, Shahzad},
  journal={Journal of Electronic Imaging},
  volume={31},
  number={5},
  pages={053010--053010},
  year={2022},
  publisher={Society of Photo-Optical Instrumentation Engineers}
}

@article{salmona2022deoldify,
  title={Deoldify: A review and implementation of an automatic colorization method},
  author={Salmona, Antoine and Bouza, Luc{\'\i}a and Delon, Julie},
  journal={Image Processing On Line},
  volume={12},
  pages={347--368},
  year={2022}
}

@inproceedings{zabari2023diffusing,
  title={Diffusing colors: Image colorization with text guided diffusion},
  author={Zabari, Nir and Azulay, Aharon and Gorkor, Alexey and Halperin, Tavi and Fried, Ohad},
  booktitle={SIGGRAPH Asia 2023 Conference Papers},
  pages={1--11},
  year={2023}
}

@article{liang2025control,
  title={Control Color: Multimodal Diffusion-Based Interactive Image Colorization: Z. Liang et al.},
  author={Liang, Zhexin and Li, Zhaochen and Zhou, Shangchen and Li, Chongyi and Loy, Chen Change},
  journal={International Journal of Computer Vision},
  volume={133},
  number={11},
  pages={7897--7923},
  year={2025},
  publisher={Springer}
}

@INPROCEEDINGS{7780459,
  author={He, Kaiming and Zhang, Xiangyu and Ren, Shaoqing and Sun, Jian},
  booktitle={2016 IEEE Conference on Computer Vision and Pattern Recognition (CVPR)}, 
  title={Deep Residual Learning for Image Recognition}, 
  year={2016},
  volume={},
  number={},
  pages={770-778},
  doi={10.1109/CVPR.2016.90}
}

@INPROCEEDINGS{10222021,
  author={Zhang, Zicheng and Li, Chunyi and Sun, Wei and Liu, Xiaohong and Min, Xiongkuo and Zhai, Guangtao},
  booktitle={2023 IEEE International Conference on Multimedia and Expo Workshops (ICMEW)}, 
  title={A Perceptual Quality Assessment Exploration for AIGC Images}, 
  year={2023},
  volume={},
  number={},
  pages={440-445},
  doi={10.1109/ICMEW59549.2023.00082}
  }

@article{li2023agiqa,
  title={Agiqa-3k: An open database for ai-generated image quality assessment},
  author={Li, Chunyi and Zhang, Zicheng and Wu, Haoning and Sun, Wei and Min, Xiongkuo and Liu, Xiaohong and Zhai, Guangtao and Lin, Weisi},
  journal={IEEE Transactions on Circuits and Systems for Video Technology},
  volume={34},
  number={8},
  pages={6833--6846},
  year={2023},
  publisher={IEEE}
}

@inproceedings{fang2020cvpr,
title={Perceptual Quality Assessment of Smartphone Photography},
author={Fang, Yuming and Zhu, Hanwei and Zeng, Yan and Ma, Kede and Wang, Zhou},
booktitle={IEEE Conference on Computer Vision and Pattern Recognition},
pages={3677-3686},
year={2020}
}

@inproceedings{li2024coco,
  title={Coco-lc: Colorfulness controllable language-based colorization},
  author={Li, Yifan and Bai, Yuhang and Yang, Shuai and Liu, Jiaying},
  booktitle={Proceedings of the 32nd ACM International Conference on Multimedia},
  pages={10939--10947},
  year={2024}
}

@inproceedings{du2024multicolor,
  title={Multicolor: Image colorization by learning from multiple color spaces},
  author={Du, Xiangcheng and Zhou, Zhao and Wu, Xingjiao and Wang, Yanlong and Wang, Zhuoyao and Zheng, Yingbin and Jin, Cheng},
  booktitle={Proceedings of the 32nd ACM International Conference on Multimedia},
  pages={6784--6792},
  year={2024}
}

@article{chen2024topiq,
  title={Topiq: A top-down approach from semantics to distortions for image quality assessment},
  author={Chen, Chaofeng and Mo, Jiadi and Hou, Jingwen and Wu, Haoning and Liao, Liang and Sun, Wenxiu and Yan, Qiong and Lin, Weisi},
  journal={IEEE Transactions on Image Processing},
  volume={33},
  pages={2404--2418},
  year={2024},
  publisher={IEEE}
}

@article{bai2025qwen3,
  title={Qwen3-vl technical report},
  author={Bai, Shuai and Cai, Yuxuan and Chen, Ruizhe and Chen, Keqin and Chen, Xionghui and Cheng, Zesen and Deng, Lianghao and Ding, Wei and Gao, Chang and Ge, Chunjiang and others},
  journal={arXiv preprint arXiv:2511.21631},
  year={2025}
}

\end{document}